\documentclass[letterpaper]{article} % DO NOT CHANGE THIS
\usepackage{aaai24}  % DO NOT CHANGE THIS
\usepackage{times}  % DO NOT CHANGE THIS
\usepackage{helvet}  % DO NOT CHANGE THIS
\usepackage{courier}  % DO NOT CHANGE THIS
\usepackage[hyphens]{url}  % DO NOT CHANGE THIS
\usepackage{graphicx} % DO NOT CHANGE THIS
\usepackage{natbib}  % DO NOT CHANGE THIS AND DO NOT ADD ANY OPTIONS TO IT
\usepackage{caption} % DO NOT CHANGE THIS AND DO NOT ADD ANY OPTIONS TO IT
\usepackage{algorithm}
\usepackage{algorithmic}

\usepackage{newfloat}
\usepackage{listings}
\DeclareCaptionStyle{ruled}{labelfont=normalfont,labelsep=colon,strut=off} % DO NOT CHANGE THIS
\floatstyle{ruled}
\newfloat{listing}{tb}{lst}{}
\floatname{listing}{Listing}
\usepackage{algorithm}
\usepackage{algorithmic}
\usepackage{multirow}
\usepackage{graphicx}
\usepackage{subfig}
\usepackage{placeins}
\usepackage{amsmath}
\usepackage{amsfonts}
\usepackage{booktabs}
\usepackage{amsthm, amssymb}

\newtheorem{theorem}{Theorem}
\newtheorem{lemma}{Lemma}

\author {
    Yichang Xu\textsuperscript{\rm 1},
    Chenwang Wu\textsuperscript{\rm 1},
    Defu Lian\textsuperscript{\rm 1}
}
\affiliations {
    \textsuperscript{\rm 1}University of Science and Technology of China\\
    xuyichang@mail.ustc.edu.cn, wcw1996@mail.ustc.edu.cn, liandefu@ustc.edu.cn
}

\title{Toward Robust Recommendation via Real-time Vicinal Defense}
\newcommand{\alg}{RVD}

\begin{document}

\maketitle

\begin{abstract}
Recommender systems have been shown to be vulnerable to poisoning attacks, where malicious data is injected into the dataset to cause the recommender system to provide biased recommendations. To defend against such attacks, various robust learning methods have been proposed. However, most methods are model-specific or attack-specific, making them lack generality, while other methods, such as adversarial training, are oriented towards evasion attacks and thus have a weak defense strength in poisoning attacks.

In this paper, we propose a general method, Real-time Vicinal Defense (\alg), which leverages neighboring training data to fine-tune the model before making a recommendation for each user. \alg \ works in the inference phase to ensure the robustness of the specific sample in real-time, so there is no need to change the model structure and training process, making it more practical. 
Extensive experimental results demonstrate that \alg \ effectively mitigates targeted poisoning attacks across various models without sacrificing accuracy. Moreover, the defensive effect can be further amplified when our method is combined with other strategies.

\end{abstract}

\section{Introduction}
With the progression of media applications, the challenge of managing copious amounts of information has become increasingly arduous for humans. To alleviate this issue, recommender systems ~\cite{10.1145/371920.372071}~\cite{5197422}~\cite{4072747}~\cite{10.1145/3397271.3401063}~\cite{Pang_2022}~\cite{sedhain2015autorec}~\cite{wu2016collaborative} have emerged. These systems utilize past user-item interactions to discern user preferences, consequently recommending items predicted to align best with an individual's taste.

Despite the convenience brought by recommender systems, current recommendation models are largely based on deep neural networks, rendering them susceptible to malicious attacks such as model inversion~\cite{10.1145/3359789.3359824}~\cite{hidano2017model}, poisoning~\cite{fang2018poisoning}~\cite{fang2020influence}, and evasive attacks~\cite{cao2017mitigating}. The ramifications of such attacks can be significant, from manipulating user preferences~\cite{fang2018poisoning} to causing complete system failure~\cite{10.1007/978-3-030-58951-6_23}. In this paper, we focus on poisoning attacks, where adversaries can inject a subset of counterfeit malicious users into the dataset to manipulate the model's behavior. Poisoning attacks are divided into untargeted attacks~\cite{DBLP:journals/corr/abs-1902-06156}~\cite{DBLP:journals/corr/abs-1911-11815}, where the attacker aims to degrade the overall performance of the recommender system, and targeted attacks~\cite{lin2020attacking}~\cite{li2016data}~\cite{fang2020influence}~\cite{lam2004shilling}, where the attacker aims to promote (or demote) certain items. For instance, within a short video recommendation system, an advertiser might register multiple malicious users who rate certain videos to boost the popularity of their advertisement videos (the target items). 

Since targeted poisoning attacks are most related to the attacker's interest, they are most likely to be abused. Therefore, it is urgent to develop an effective defense to mitigate such attacks. One direct approach is to preprocess the data before training, like~\cite{peri2020deep}~\cite{borgnia2021strong}. The disadvantage of such a defense is that only extremely abnormal data points will be filtered. However, in some more advanced attacks, like AUSH~\cite{lin2020attacking}, embedding of malicious users may be indistinguishable, and such methods will fall short in their detection, leading to data imbalance and overfitting~\cite{wang2022threats}. Some methods can only be applied to a specific model. For example, ~\cite{cheng2010robust}~\cite{bampis2017robust} are two methods based on robust training, and they are attack-agnostic in design. However, they can only be applied to matrix-factorization-based models. Another method, ~\cite{he2018adversarial}, can be applied to different models, but its initial design purpose is to mitigate evasive attacks, and it shows a weaker effect on poisoning attacks.

It is necessary to revisit attacks in recommender systems to design a defense strategy independent of attacks and robust to different models. To this end, we visualize the poisoning data generated by different attacks and find that they are often mixed among normal users. This is because the sparsity of recommendation data makes the learned embedding space also sparse, creating blind spots in high-dimensional sparse spaces where attacks can exploit vulnerabilities. This finding also demonstrates the difficulty of detecting these fake users. To address this issue, a potential approach is to use more new samples to cover more blind areas, thereby enhancing model robustness. However, considering the large number of users and items, learning a complete and continuous representation space is impractical, which also places a high demand on the expressive power of the model. To overcome these challenges, we present a Real-time Vicinal Defense (RVD), which makes a compromise and only focuses on the robustness of the current recommended users. Specifically, for each user to be predicted, RVD does not directly give a prediction because it may be affected by an attack. Instead, it is first trained based on neighborhood data to eliminate the influence of poisoned data on this user and then make predictions. Obviously, RVD only focuses on the inference stage and does not modify the model architecture or training mode, which makes it versatile and could be applied to various recommendation models. Besides, by adopting this approach, it is not necessary to guarantee the robustness of all regions, reducing the demands placed on the model. In the inference phase, the model undergoes an unlearning process, where the influence of malicious users is unlearned. In our theoretical part, we prove that the unlearning strength is guaranteed.
Our experiments across four datasets demonstrate that our method significantly mitigates targeted poisoning attacks compared to baselines. Moreover, we illustrate that by integrating our defense mechanism into existing defense methods, we can further enhance their performance.

Our main contributions are summarized as follows:

\begin{itemize}
  \item We propose Real-time Vicinal Defense (\alg) to bolster the robustness of a model. It utilizes vicinal training data during the inference phase to fine-tune the model. This is an unlearning procedure, and we theoretically guarantee its unlearning strength.
  \item Our approach has been demonstrated to effectively mitigate targeted poisoning attacks across various models without sacrificing accuracy. Moreover, the defensive effect can be further amplified when our method is combined with other strategies.
  \item Our method only focuses on the inference phase, which means that the model architecture and the training mode will not be altered. This makes our method more general and practicable.
\end{itemize}

\section{Related Work}

\subsection{Poisoning Attacks}

The principal focus of our study centers around a unique variety of adversarial manipulation known as poisoning attacks. This form of offensive strategy in the realm of recommender systems has garnered increased scholarly attention in recent years, leading to the proposal and development of a multitude of poisoning attacks.

The evolution of poisoning attacks started with Lam and Riedl's pioneering work~\cite{lam2004shilling}. They introduced two simple yet effective methods: RANDOM and AVERAGE. Next in the timeline came the concept of the PGA attack, proposed by Li et al.~\cite{li2016data}, which involves the application of Stochastic Gradient Descent (SGD) to optimize a specific poisoning-focused objective function, subtly changing the ratings of the top-rated filler items. Following this, Lin et al.~\cite{lin2020attacking} contributed to an innovative AUSH strategy. By using Generative Adversarial Networks (GANs), they devised a way to create 'imperceptible' synthetic users capable of assigning the highest possible rating to a target item. This trend was furthered by Fang et al.~\cite{fang2020influence} with their TNA attack, where they used the most influential users in the system to increase the ratings of targeted items. Most recently, ~\cite{10122715} put forth the MixInf and MixRand methods. These integrate a threat estimator and a user generator to create reasonable yet malicious users capable of poisoning the system, requiring no model retraining. This sequence of developments not only underscores the escalation of sophistication in poisoning attack strategies but also emphasizes the critical need for the corresponding evolution in defensive mechanisms.

\subsection{Defense against Poisoning Attacks}
Such defenses can be divided into two types: data preprocess and robust training. In data preprocess-based approaches, malicious users are expected to be filtered out. For example, an outlier-based filter removes data points situated far from the dataset centroid~\cite{steinhardt2017certified}. Such methods offer a level of robustness when malicious users look distinguishable. However, its effectiveness may diminish if malicious data points blend into the boundary around the centroid. Another approach proposed by ~\cite{peri2020deep} aggregates all close neighbors of each data point with the same label into the new training dataset. While this defense may also be subverted by modern attacks, it is also inapplicable for matrix-completion-based recommender systems. Instead of filtering the training data, some defenses perform data augmentation~\cite{zhang2017mixup}. Such methods will increase the sampling complexity for even a slightly large dataset and they are less suitable for matrix-completion-based recommender systems because items with a rating of 0 do not necessarily denote a low rating, and the application of convex combinations to rating vectors could result in loss of sparsity.

The most applicable defense is based on robust training. Such methods alter the training procedure to make the model unaware of malicious users. He et al.~\cite{he2018adversarial} proposed a robust technique known as Adversarial Personalized Ranking (APR), which optimizes the model through a minimax game. Cheng et al.~\cite{cheng2010robust} proposed the Least Trimmed Square Matrix Factorization (LTSMF) method. This approach leverages a robust loss function, disregarding user-item pairs for which the loss falls within the top-largest $k$ percent, thereby enhancing system robustness. More recently, Bampis et al.~\cite{bampis2017robust} introduced Robust Matrix Factorization (RMF), a unique technique that employs a robust optimization target to make the system less susceptible to manipulations. Such attacks are generally stronger than data preprocessing-based approaches because most of them are not attack-agnostic. However, they are focusing on a specific model, like matrix factorization. We need to further develop a defense that can be applied in various models and various attacks.

\section{Problem Definition}
% We use explicit feedback in our experiments, which means that each rating is an integer and 0 indicates no interaction. The recommender system predicts the scores for uninteracted items and recommends items with top-k predicted scores.

{\bfseries Attacker's Goal}. Poisoning attacks can be divided into targeted attacks and untargeted attacks. In untargeted attacks, the attacker aims to degrade the model's overall performance. Targeted attacks can further be divided into promotion attacks and demotion attacks, where the attacker injects malicious users to promote or demote target items. In our paper, we focus on promotion attacks. The attacker has a set of target items (5 items in our experiment) and desires those items to rank in the top $k$ highest predicted scores.

\noindent{\bfseries Attacker's Knowledge}. Given the focus of our work on defense, we shall assume the worst case of attack: the attacker knows all the data, the model structure, and the parameters. This facilitates to evaluate the upper bound of defense methods. 

\noindent{\bfseries Attacker's Capability}. Since injecting a large number of users is hard and is easy to be detected, in our work, the attacker injects about 3\% fake users into the original dataset. Moreover, since the anomaly detection mechanism will also be triggered if a user rates too many items, we limit the number of ratings of each fake user to the average number of ratings per user.

\noindent{\bfseries Defender's Knowledge}. We assume that the defender is the system owner: he can access the entire dataset and knows the model's structure and hyperparameters. However, he is attack-unaware, so the attacking algorithm and targeted items are unknown.

\noindent{\bfseries Defense Goal}. Against promotion attacks, a good defense strategy should: (1) keep the popularity of the attacked items the same as when it was not attacked; (2) maintain recommendation accuracy. The key recommendation performance cannot be neglected due to excessive focus on model robustness; (3) The extra time increase due to defense should be tolerable, which is especially critical in real-time recommendations.

\section{Methodology}
\subsection{Analysis of Poisoning Attacks}
\begin{figure}
	\centering
    \subfloat[PGA]
	{\includegraphics[width=0.25 \textwidth]{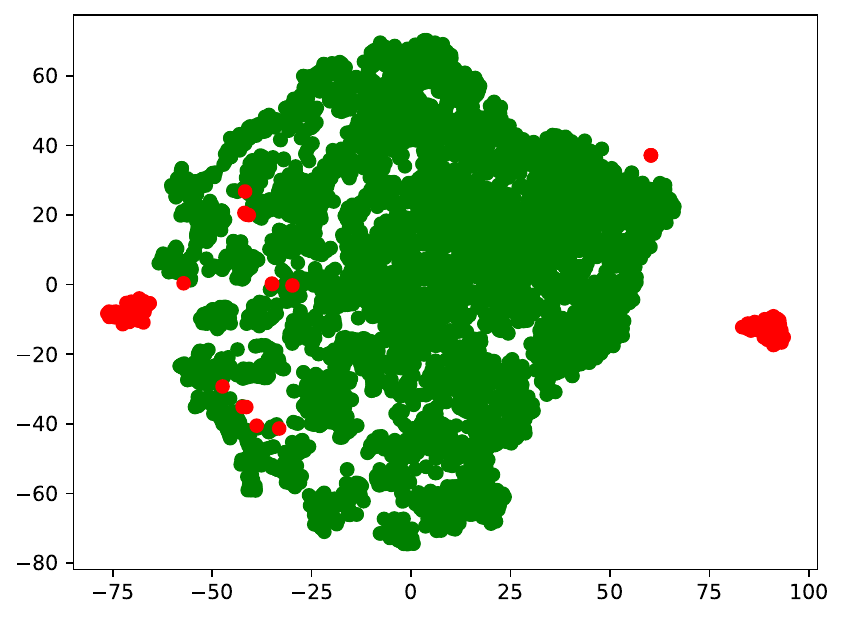}}
	\subfloat[MixInf]
	{\includegraphics[width=0.25 \textwidth]{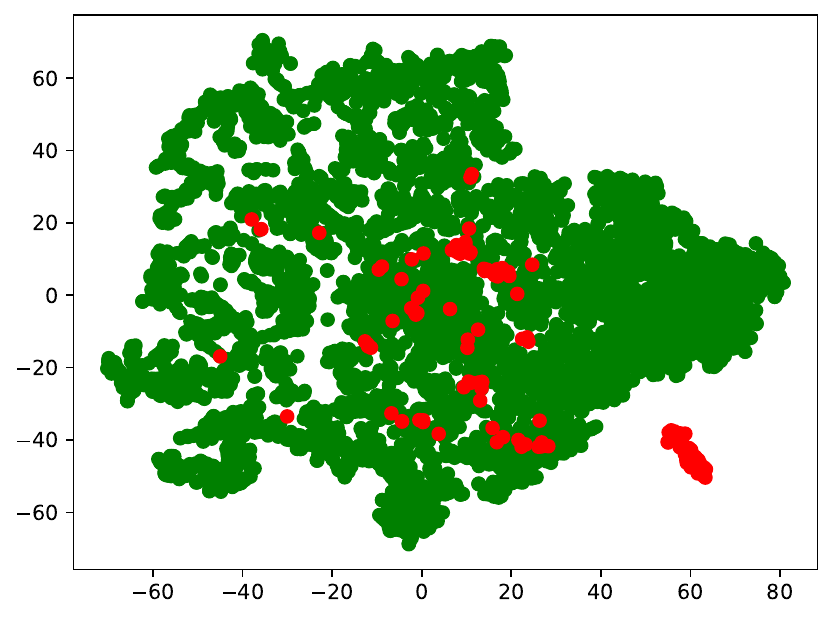}}
 \caption{TSNE visualization of poisoning data generated by PGA and MixInf. Red: poisoning data. Green: real data.}
\label{tsne}
\end{figure}

To design a general defense strategy, it is necessary to explore the common characteristics of different attacks, especially those that simple defense strategies cannot mitigate. To this end, on the ML-1M dataset, We choose two attacks, PGA~\cite{li2016data} and MixInf~\cite{10122715}, to generate poisoning data for attacking AutoRec~\cite{sedhain2015autorec}. We then extract each data point's latent representation (hidden features of the model) and use TSNE~\cite{JMLR:v9:vandermaaten08a} to visualize them. The resulting plots are shown in Figure \ref{tsne}(a) and Figure \ref{tsne}(b).

From those results, we find that the malicious users generated by the attack are often mixed with the benign users. This may be due to the sparsity of the data (each user only has very few ratings), resulting in discrete high-dimensional representations learned by the model. Then the unknown blind zone model cannot guarantee accurate inference, leading to the possible existence of poisoned data. In addition, this indistinguishability also brings difficulties to detection. This is because existing methods often exclude normal users from pursuing a high detection rate, which is unreasonable.

\subsection{Real-time Vicinal Defense}
According to the findings in the previous subsection, a possible way to improve the recommendation robustness is to allow the model learning to cover these blind areas as much as possible. A naive way is to use data augmentation so that augmented data will cover most of those blind spots. However, this is impractical because the embedding space is sparse, and covering all those blind spots becomes extremely hard. In addition, even with the rich augmentation data support, the model needs to provide enough expressiveness to learn all the data points, which is also extremely demanding on the model.

Therefore, we compromise: we do not require the model to be robust for all users but only for the specific user we infer. Naturally, we can fine-tune the user's neighborhood representation before making specific recommendations.
At this time, given the current small neighborhood data for model training, its learning pressure will be significantly reduced. This may help it better learn the neighborhood representation to cover as many learning blind spots around the user as possible that are easily overlooked. 
Based on these considerations, we propose a Real-time Vicinal Defense (\alg) that forces the model to learn a robust neighborhood by learning the user neighborhood data before giving recommendations for each user. It is undeniable that using neighborhood data may include malicious data, but considering that malicious data is already rare in the entire set, it will be even rarer in the neighbors.

Specifically, our defense is applied in the inference phase of the model, so the recommendation model is trained on $n$ users (including $n'$ fake users where $n’\ge 0$) as usual. In the inference stage, for user $u$, since we do not know whether he has been attacked, we do not recommend it directly but retrain its neighborhood data to reduce possible harm. So, first, we need to get neighborhood data for each user. Specifically, suppose the embedding vector for user $u$ is $\mathbf{z}_u$, then we first calculate the embedding distance between user $u$ and each user, i.e.,
$$
d_{iu}=\|\mathbf{z}_i-\mathbf{z}_u\|_2,i=i,2,\cdots n.
$$
Then we choose $k$ users whose distance $d_{iu}$ is the smallest. Denote this subset of users as $\mathcal{D}'$. Finally, we train the model on $\mathcal{D}'$ to unlearn the influence of malicious users. We expect this unlearning step can help to reduce the impact of attacks on user $u$. After vicinal training, we give the recommendation result to user $u$ based on the fine-tuned model.

The formalized algorithm can be represented in Algorithm \ref{alg1}. Line 4 trains the base model on the poisoned dataset. Line 5 stores the parameter of the base model so that it can be recovered before giving a prediction for each new user. Line 7-9 computes the distance between the latent embedding of each user in the dataset and that of the target user $u$. Line 10 sorts the distances and line 11 choose $k$ neighbors to form a new subset $\mathcal{D}'$. Line 12 fine-tunes the base model on newly-formed $\mathcal{D}'$ and line 13 makes the predicted scores on all items of the target user. Finally, line 14 recovers the base model's parameter to give a prediction for the next target user.
\begin{algorithm}[t]
  \caption{Real-time Vicinal Defense}
  \label{alg1}
  \begin{algorithmic}
    \renewcommand{\algorithmicrequire}{\textbf{Input}}
    \REQUIRE base model $M$, poisoned dataset $\mathcal{D}$, number of chosen neighbors $k$, $n$ users' latent embedding $\mathbf{z}_1,\mathbf{z}_2,\cdots,\mathbf{z}_n$, model parameter $\mathbf{\Theta}$.
    \STATE train $M$ on $\mathcal{D}$
    \STATE $\mathbf{\Theta}'\leftarrow\mathbf{\Theta}$
    \FOR{$u$ in $\mathcal{D}$}
        \FOR{$i$ in $\mathcal{D}$}
            \STATE $d_{iu}\leftarrow \|\mathbf{z}_u-\mathbf{z}_i\|_2$
        \ENDFOR
        \STATE $\{d_{i_1u}, d_{i_2u}, \cdots, d_{i_nu}\}\leftarrow sort(\{d_{1u}, d_{2u}, \cdots d_{nu}\})$
        \STATE $\mathcal{D'}\leftarrow\{i_1,i_2,\cdots,i_k\}$
        \STATE train $M$ on $\mathcal{D}'$
        \STATE Yield predicted scores for $u$ as $M(u)$
        \STATE $\mathbf{\Theta}\leftarrow\mathbf{\Theta}'$
    \ENDFOR
  \end{algorithmic}
\end{algorithm}

\noindent\textbf{Efficiency Analysis.}
For each user, the time complexity of selecting $k$ nearest neighbors is $O(n\log n+ne)$, which can be approximated to $O(n\log n)$, given that $n$ is typically much larger than $e$, where $e$ is the dimension of the embedding. The total time complexity for providing robust recommendations for one user is $O(f(k) + n\log n + g)$. Here, $g$ represents the time complexity of generating predicted scores for each user, and $f(k)$ is the time complexity of fine-tuning the network using $k$ neighbors. In the experiment, we will demonstrate that this inference delay is imperceptible for each user.

We also provide some potential approaches to improve efficiency as future research:
\begin{itemize}
    \item For matrix-factorization-based models, we can leverage some optimizations like the K-D tree to accelerate the KNN algorithm. For example, for K-D tree, the time complexity of selecting $k$ nearest neighbors could be reduced to $O(e\log n)$. For deep-learning-based models, we can first calculate $k$ neighbors for each user in the training set. For each user in the test set, we can find the closest user in the training set and use its $k$ neighbors as an approximation. In this case, after computing $k$ neighbors for each user in the training set, in the inference phase, it only takes $O(e)$ to get $k$ neighbors for a single user.
    \item Each user's representation does not change drastically, so we do not need to calculate its neighbors every time it requests a recommendation. We can cache the result and update it periodically. In this way, the search for nearest neighbors is avoided.
    \item For the small model, we could distribute the fine-tuning work to the client to relieve the pressure on the server. In this case, the server can deliver the model and the ratings of $k$ nearest users to the client, and the client fine-tunes the model in its local platform. Since $k$ is usually small, it is easy to fine-tune the network locally without too much hardware requirement.
\end{itemize}

% Note that such method may rise a privacy concern in some situation because each client knows ratings of its neighboring users. To further address this issue, the server can shuffle items before sending the model and neighboring users to a client. First, for items $\{1,2,\cdots,m\}$, suppose the predicted rating for item $i$ of 
% user $u$ is given by $\hat{y}_u^i=f_{\Theta; \theta_i}(u)$. $\Theta$ is the common parameter useful in all items. For matrix factorization case, $\Theta$ is the user embedding. 
% $\theta_i$ is the parameter used specially for item $i$, like the $i$th item embedding in matrix factorization model. Before sending the ratings and the model to the user, we first shuffle the items to be like 
% $I=\{i_1,\cdots,i_m\}$. Then, the network parameters are modified as $\{\theta_1,\cdots,\theta_m\}\leftarrow\{\theta_{i_i},\cdots,\theta_{i_m}\}$. The shuffling work is done in the server and is transparent to clients. Each client only need to fine-tune the model just like items are not shuffled. The client can only receive the shuffled ratings for a little part of users, and it cannot 
% know who those users are and what those items are. Of course, if that client modifies his ratings thousands of times in a short period and compare the returned users together, it may be able to recognize the items. 
% We can maximally avoid this by limiting the number of requests and randomly invoke some certified users into the chosen neighbors.
\subsection{Theoretical Bound}
Our algorithm chooses the neighbors of each user to fine-tune the model. This is based on an observation that most users' neighbors are benign users. Intuitively, the influence of malicious users is unlearned during the fine-tuning phase. In this section, we theoretically prove that the unlearning abilities are certified. In other words, after fine-tuning the model, the distance between the current and optimal parameters is within a certain bound. 
% Our proof is motivated by paper~\cite{neel2020descenttodelete}.
% We first give the following definitions for strong convexity, Lipschitzness and smoothness.

% \begin{definition}
% \label{def1}
% (Strong Convexity) A function $f:\Theta\to\mathbb{R}$ is m-strongly convex if
% $$
% \exists m\ge0, \forall \theta_1,\theta_2\in\Theta, 0<t<1,
% $$
% $$
% f(t\theta_1+(1-t)\theta_2)\le tf(\theta_1)+(1-t)f(\theta_2)-\frac{m}{2}t(1-t)\|\theta_1-\theta_2\|_2^2.
% $$
% \end{definition}

% \begin{definition}
% \label{def2}
% (Lipschitzness) A function $f:\Theta\to\mathbb{R}$ is L-Lipschitz if
% $$
% \forall \theta_1,\theta_2\in\Theta,
% $$
% $$
% |f(\theta_1)-f(\theta_2)|\le L\|\theta_1-\theta_2\|_2.
% $$
% \end{definition}

% \begin{definition}
% \label{def3}
% (Smoothness) A function $f:\Theta\to\mathbb{R}$ is M-smooth if $f$ is differentiable and 
% $$
% \forall \theta_1,\theta_2\in\Theta,
% $$
% $$
% \|\nabla f(\theta_1)-\nabla f(\theta_2)\|\le M\|\theta_1-\theta_2\|_2.
% $$
% \end{definition}

% Theorem \ref{theorem2} is proposed in~\cite{ABD17}.
% \begin{theorem}
% \label{theorem2}
% Let $f$ be convex and M-smooth, $\theta^{*}=\text{argmin}_{\theta\in\Theta}f(\theta)$. After $T$ steps of gradient descent with step size $\eta=\frac{1}{M}$, 
% $$
% f(\theta_T)-f(\theta^{*})\le\frac{M\|\theta_0-\theta^*\|_2^2}{2T}.
% $$
% \end{theorem}
% Then we can propose our first bound when the loss function is m-strongly convex, L-Lipschitz and M-smooth.
\begin{theorem}
\label{theorem2}
Suppose the dataset $\mathcal{D}_0$ has the capacity of $n$. $\mathcal{D}_1$ is obtained by deleting $n-k$ entries from $\mathcal{D}_0$, i.e., $|\mathcal{D}_1|=k$. Define $\gamma=\frac{M-m}{M+m}$. Loss function $f:\Theta\to\mathbb{R}$ is m-strongly convex, L-Lipschitz, and M-smooth on any dataset. After running the gradient descent algorithm on $\mathcal{D}_0$ for $T$ steps with learning rate $\eta=\frac{2}{M+m}$, where $T\ge I-\log_{\gamma}\left(\frac{Dmn}{2L}\right)$ and $D$ is the diameter of $\Theta$, and then running the gradient descent algorithm on $\mathcal{D}_1$ with the same learning rate for $I$ steps and get the parameter $\hat{\theta}_{\mathcal{D}_1}$, we have
$$
|f_{\mathcal{D}_1}(\hat{\theta}_{\mathcal{D}_1})-f_{\mathcal{D}_1}(\theta_{\mathcal{D}_1}^{*})|=O\left(\frac{L^2\gamma^{2I}}{m^2M}\log^2\left(\frac{n}{k}\right)\right),
$$
where $\theta_{\mathcal{D}_1}^{*}=\arg\min_{\theta\in\Theta}f_{\mathcal{D}_1}(\theta)$
\end{theorem}

\begin{theorem}
\label{theorem3}
Suppose the dataset $\mathcal{D}_0$ has the capacity of $n$. $\mathcal{D}_1$ is obtained by deleting $n-k$ entries from $\mathcal{D}_0$, i.e., $|\mathcal{D}_1|=k$. Define $\gamma=\frac{M}{M+2m}$. Loss function $f:\Theta\to\mathbb{R}$ is convex, L-Lipschitz, and M-smooth on any dataset. After running the gradient descent algorithm to minimize the regularized $f$ (defined as $g(\theta)=f(\theta)+\frac{m}{2}\|\theta\|_2^2$) on $\mathcal{D}_0$ for $T$ steps with learning rate $\eta=\frac{2}{M+2m}$, where $T\ge I-\log_{\gamma}\left(\frac{Dmn}{2L}\right)$ and $D$ is the diameter of $\Theta$, and then running the gradient descent algorithm on $\mathcal{D}_1$ with the same learning rate for $I$ steps and get the parameter $\hat{\theta}_{\mathcal{D}_1}$, we have
\begin{align*}
    &|f_{\mathcal{D}_1}(\hat{\theta}_{\mathcal{D}_1})-f_{\mathcal{D}_1}(\theta_{\mathcal{D}_1}^{*})| \\
&=O\left(\frac{(L+mD)^2\gamma^{2I}}{m^2(M+m)}\log^2\left(\frac{n}{k}\right)+mD^2\right),
\end{align*}
where $\theta_{\mathcal{D}_1}^{*}=\arg\min_{\theta\in\Theta}f_{\mathcal{D}_1}(\theta)$
\end{theorem}

These theorems are based on the assumption that the loss function is strongly convex or convex. This is reasonable because the recommendation model is usually designed to have a convex loss function to avoid falling into a local optimum.
We can see from Theorem \ref{theorem2} and Theorem \ref{theorem3} that for a larger $k$, it is easier to unlearn the malicious influence. Also, the generality of the model will be close to that of the originally trained base model or be even better since there are fewer malicious data points. However, for a larger $k$, malicious points may not be sufficiently filtered. In our experiments, it is important to find an appropriate $k$. The proofs of Theorem \ref{theorem2} and Theorem \ref{theorem3} are shown in Appendix A and impacts on different $k$ are shown in Appendix E.

\section{Experiments}
\subsection{Experimental Setup}
\subsubsection{Datasets.}
We conduct experiments on four datasets widely adopted by existing work \cite{10122715,fang2020influence}: ML-100K~\cite{movielens100k}, ML-1M~\cite{movielens1m}, FilmTrust~\cite{1593032}, and Yelp~\cite{yelp}. ML-100K is a classic dataset with ratings from 943 users for 1682 films. ML-1M collects ratings from 6040 users across 3706 films. FilmTrust consists of ratings from 796 users on 2011 films, with a rating scale from 1 to 8. Yelp includes ratings from 14575 users for 25602 items and offers a scale from 1 to 5. 
The detailed statistics of the four datasets are shown in table \ref{tab:statistics}. Due to space limitations, by default, the main text shows the performance on ML-1M and Yelp, and the other two datasets are placed in the appendix.
\begin{table}
\centering
\renewcommand\arraystretch{0.85}
  \begin{tabular}{l|c|c|c}
    \midrule
    Dataset & Users & Items & Ratings\\
    \midrule
    ML-100K & 943 & 1682 & 100000 \\
    ML-1M & 6040 & 3706 & 1000209 \\
    FilmTrust & 796 & 2011 & 30880 \\
    Yelp & 14575 & 25602 & 555374 \\
    \midrule
  \end{tabular}
  \caption{Statistics of Datasets}
  \label{tab:statistics}
\end{table}
\subsubsection{Poisoning Attacks.}
We consider seven distinct types of poisoning attacks. These attacks include AUSH~\cite{lin2020attacking}, which utilizes Generative Adversarial Networks to generate imperceptible users who assign the highest possible rating to a target item. PGA~\cite{li2016data} involves manipulating the ratings of selected top-rated items to subtly poison the system. TNA~\cite{fang2020influence} focuses on identifying influential users and leveraging them to assign maximum ratings to targeted items. MixInf~\cite{10122715} combines an influence-based threat estimator and a user generator to create malicious users for poisoning. MixRand~\cite{10122715} is a non-influential version of MixInf that directly assigns the maximum rating to target items. RANDOM~\cite{lam2004shilling} assigns the maximum rating to target items and randomly selects filler items based on a distribution mimicking the overall dataset. AVERAGE~\cite{lam2004shilling} follows a similar approach but assigns completely random ratings to filler items. Technical details are given in Appendix B\ref{sec:appB}. Each attack target promotes five random non-popular items because promoting these items is more practical to the attacker.

% A comprehensive account of the number of malicious users injected by each attack into each dataset is outlined in Table \ref{tab:injected}. 

% \begin{table}
% \centering
%   \caption{Number of Malicious Users Injected Per Dataset}
%   \label{tab:injected}
%   \begin{tabular}{|l|c|}
%   \hline
%     Dataset & \#Malicious Users\\
%     \hline
%     ML-100K & 28 \\
%     ML-1M & 181 \\
%     FilmTrust & 23 \\
%     Yelp & 473 \\
%     \hline
%   \end{tabular}
% \end{table}

\begin{table}
\centering
\renewcommand\arraystretch{0.85}
\begin{tabular}{l|c|c|c|c}
\midrule
Dataset & AutoRec & CDAE & MF & NSVD \\ \midrule
ML-100K & 12 & 12 & 12 & 25 \\ 
ML-1M & 12 & 12 & 12 & 70 \\ 
FilmTrust & 12 & 12 & 650 & 7 \\
Yelp & 100 & 200 & 100 & 900 \\ \midrule
\end{tabular}
\caption{Choice of $k$ in different base models and datasets.}
\label{tab:param}
\end{table}

\begin{table*}[htbp]
  \centering
  \footnotesize
  \renewcommand\arraystretch{0.85}
  \setlength{\tabcolsep}{0.006\linewidth}{
    \begin{tabular}{c|c|c|c|c|c|c|c|c|c|c|c|c|c|c}
    \toprule
    \multicolumn{1}{c|}{Dataset} & \multicolumn{7}{c|}{ML-1M (HR@50)}                            & \multicolumn{7}{c}{Yelp (HR@500)} \\
    \midrule
    Attack & AUSH  & PGA   & TNA   & MixInf & MixRand & Random & Average & AUSH  & PGA   & TNA   & MixInf & MixRand & Random & Average \\
    \midrule
    no defense & 0.999 & 0.928 & 0.998 & 1.000 & 1.000 & 0.976 & 0.998 & 0.836 & 0.954 & 0.982 & 0.965 & 0.975 & 0.975 & 0.163 \\
    APR   & 0.596 & 0.453 & 0.462 & 0.995 & 0.773 & 0.029 & 0.501 & 0.990 & 0.982 & 0.569 & 0.998 & 0.989 & 0.983 & 0.978 \\
    LTSMF & 0.400 & 0.197 & 0.799 & 0.200 & 1.000 & 0.671 & 0.599 & 0.921 & 0.553 & 0.966 & 0.937 & 0.971 & 0.009 & 0.011 \\
    RMF   & 0.064 & 0.214 & 0.166 & \textbf{0.027} & 0.075 & 0.212 & 0.199 & 0.502 & 0.499 & 0.516 & 0.542 & 0.575 & 0.410 & 0.415 \\
    Our   & \textbf{0.011} & \textbf{0.004} & \textbf{0.002} & 0.564 & \textbf{0.042} & \textbf{0.004} & \textbf{0.001} & \textbf{0.162} & \textbf{0.148} & \textbf{0.170} & \textbf{0.198} & \textbf{0.170} & \textbf{0.185} & \textbf{0.179} \\
    \bottomrule
    \end{tabular}%
    }
    \caption{Target items' hit ratios after different defenses on ML-1M and Yelp. Notably, their hit ratios are 0 when there is no attack. Therefore, the lower the hit ratio, the better the defense performance.}
  \label{tab:comp_sample}%
\end{table*}%

\begin{table*}[htbp]
  \centering
  \footnotesize
  \renewcommand\arraystretch{0.85}
  \setlength{\tabcolsep}{0.0052\linewidth}{
    \begin{tabular}{c|c|c|c|c|c|c|c|c|c|c|c|c|c|c|c}
    \toprule
    \multicolumn{2}{c|}{Dataset} & \multicolumn{7}{c|}{ML-1M (HR@50)}                            & \multicolumn{7}{c}{Yelp (HR@500)} \\
    \midrule
    Model & Attack & AUSH  & PGA   & TNA   & MixInf & MixRand & Random & Average & AUSH  & PGA   & TNA   & MixInf & MixRand & Random & Average \\
    \midrule
    \multirow{2}[2]{*}{AutoRec} & before & 0.611 & 0.152 & 0.853 & 0.806 & 0.229 & 0.579 & 0.774 & 1.000 & 1.000 & 1.000 & 1.000 & 1.000 & 1.000 & 1.000 \\
          & after & 0.004 & 0.001 & 0.007 & 0.016 & 0.038 & 0.000 & 0.008 & 0.002 & 0.001 & 0.002 & 0.001 & 0.001 & 0.001 & 0.001 \\
    \midrule
    \multirow{2}[2]{*}{CDAE} & before & 1.000 & 1.000 & 1.000 & 1.000 & 1.000 & 1.000 & 1.000 & 1.000 & 1.000 & 1.000 & 1.000 & 1.000 & 1.000 & 1.000 \\
          & after & 0.003 & 0.001 & 0.002 & 0.128 & 0.074 & 0.001 & 0.003 & 0.288 & 0.411 & 0.167 & 0.919 & 0.810 & 0.340 & 0.370 \\
    % \midrule
    % \multirow{2}[2]{*}{MF} & before & 0.999 & 0.928 & 0.998 & 1.000 & 1.000 & 0.976 & 0.998 & 0.836 & 0.954 & 0.982 & 0.965 & 0.975 & 0.975 & 0.163 \\
    %       & after & 0.011 & 0.004 & 0.002 & 0.564 & 0.042 & 0.004 & 0.001 & 0.168 & 0.018 & 0.095 & 0.068 & 0.174 & 0.024 & 0.024 \\
    \midrule
    \multirow{2}[2]{*}{NSVD} & before & 0.093 & 0.091 & 0.079 & 0.836 & 0.898 & 0.040 & 0.094 & 0.731 & 0.720 & 0.023 & 0.777 & 0.883 & 0.596 & 0.850 \\
          & after & 0.004 & 0.029 & 0.007 & 0.252 & 0.334 & 0.005 & 0.008 & 0.156 & 0.127 & 0.003 & 0.253 & 0.743 & 0.092 & 0.331 \\
    \bottomrule
    \end{tabular}%
    }
    \caption{Target items' hit ratios under different recommendation models on ML-1M and Yelp. Their hit ratios are 0 when there's no attack. Therefore, the lower the hit ratio, the better the defense performance.}
  \label{tab:hr_sample}%
\end{table*}%

\subsubsection{Defense Baselines.}
The selected baselines for our experiments are as follows: Adversarial Personalized Ranking (APR)~\cite{he2018adversarial} optimizes the model through a minimax game, where the loss function includes a term for the adversarial perturbation. Least Trimmed Square Matrix Factorization (LTSMF)~\cite{cheng2010robust} uses a robust loss function and trims the top largest percentage of $(u,i)$ pairs. Robust Matrix Factorization (RMF)~\cite{bampis2017robust} focuses on minimizing the Frobenius norm between the user-item rating matrix and the reconstructed matrix using a specific operator. The detailed descriptions of those baselines are provided in Appendix C\ref{sec:appC}.

\subsubsection{Evaluation Metrics.}
To assess the effectiveness and robustness, we employ two primary metrics: Accuracy and hit ratio (HR@K). We split the dataset using the leave-one-out method. The accuracy is calculated by dividing the sum of indicator functions, which check if such a test item is recommended to the user, by the total number of users whose actual rating meets the threshold. The hit ratio (HR@K) measures how often targeted items in a poisoning attack are recommended among the top-K recommendations. The results shown are the average HR of all target items. By default, $K$ is set to 50 in the first three data sets, and Yelp is set to 500.
Detailed formulas calculating the accuracy and the hit ratio are shown in Appendix D\ref{sec:appD}.
\subsubsection{Parameter Settings.}

The training of RVD is consistent with the original model, and the only additional parameter is $k$, representing the number of neighbors selected for fine-tuning the base model. For a fair comparison, we maintain a consistent $k$ across all poisoning attacks for each model on each dataset. However, as the underlying data distribution and the number of benign and malicious users vary across different datasets and models, an optimal $k$ can differ substantially.
Table \ref{tab:param} presents the chosen values of $k$ in our experiment corresponding to different datasets and base models. We have also conducted a sensitivity analysis w.r.t. $k$ in Appendix E.

\subsection{Defense Performance Evaluation}
\subsubsection{Comparison with Baselines.}
Constrained by defense baselines are specially designed for the matrix factorization model, we experiment with it as the target model, and the results are shown in Table \ref{tab:comp_sample}.
Our method significantly outperforms the baselines in most cases. As an extreme case, we can see when ML-1M is attacked using TNA, the hit ratio rises to 0.998. After the defense of the strongest RMF, it amazingly drops to 0.002 -- nearly 0. We can see for ML-1M, our defense can drag the hit ratio to below 0.005 in most cases, and for Yelp, the hit ratio drops to around 0.17 after using our method. This is much better compared with baselines.

\subsubsection{Performance in Other Recommendation Models.}
As we emphasized earlier, our approach is general because it requires no changes to the model or training schema. This section evaluates the defense capabilities of RVD on other proposed models. Here we choose AutoRec~\cite{sedhain2015autorec}, CDAE~\cite{wu2016collaborative}, and NSVD~\cite{paterek2007improving} for experiment.
The result on ML-100K is shown in Table~\ref{tab:hr_sample}, and additional results on other datasets are shown in Appendix E. It can be seen that under various recommendation models, our method can significantly improve the recommendation robustness.

Interestingly, deep-learning-based models (AutoRec and CDAE) are generally easier to defend. For instance, when the model trained on ML-1M is poisoned by MixInf, from Table \ref{tab:hr_sample}, we can find that for MF and NSVD, the hit ratio after defense becomes 0.564 and 0.252, respectively. However, in AutoRec, it drops to 0.016, while in CDAE, it drops to 0.128. One possible reason is that deep learning based models can better learn representations of users.

\begin{table*}[htbp]
  \centering
  \footnotesize
  \renewcommand\arraystretch{0.85}
  \setlength{\tabcolsep}{0.0052\linewidth}{
    \begin{tabular}{c|c|c|c|c|c|c|c|c|c|c|c|c|c|c|c}
    \toprule
    \multicolumn{2}{c|}{Dataset} & \multicolumn{7}{c|}{ML-1M}                            & \multicolumn{7}{c}{Yelp} \\
    \midrule
    Model & Attack & AUSH  & PGA   & TNA   & MixInf & MixRand & Random & Average & AUSH  & PGA   & TNA   & MixInf & MixRand & Random & Average \\
    \midrule
    \multirow{2}[2]{*}{AutoRec} & before & 0.058 & 0.043 & 0.061 & 0.036 & 0.022 & 0.057 & 0.055 & 0.208 & 0.207 & 0.202 & 0.207 & 0.207 & 0.204 & 0.202 \\
          & after & 0.121 & 0.115 & 0.116 & 0.118 & 0.124 & 0.117 & 0.125 & 0.232 & 0.239 & 0.238 & 0.237 & 0.239 & 0.237 & 0.236 \\
    \midrule
    \multirow{2}[2]{*}{CDAE} & before & 0.056 & 0.054 & 0.046 & 0.058 & 0.054 & 0.065 & 0.042 & 0.211 & 0.204 & 0.208 & 0.211 & 0.210 & 0.210 & 0.211 \\
          & after & 0.064 & 0.056 & 0.060 & 0.058 & 0.064 & 0.056 & 0.061 & 0.200 & 0.200 & 0.200 & 0.200 & 0.200 & 0.197 & 0.198 \\
    \midrule
    \multirow{2}[2]{*}{MF} & before & 0.072 & 0.066 & 0.063 & 0.080 & 0.079 & 0.079 & 0.068 & 0.314 & 0.313 & 0.308 & 0.322 & 0.319 & 0.300 & 0.311 \\
          & after & 0.088 & 0.097 & 0.093 & 0.085 & 0.082 & 0.088 & 0.092 & 0.325 & 0.323 & 0.323 & 0.307 & 0.320 & 0.319 & 0.326 \\
    \midrule
    \multirow{2}[2]{*}{NSVD} & before & 0.154 & 0.163 & 0.159 & 0.153 & 0.128 & 0.171 & 0.156 & 0.405 & 0.401 & 0.406 & 0.407 & 0.412 & 0.409 & 0.397 \\
          & after & 0.138 & 0.136 & 0.139 & 0.140 & 0.141 & 0.138 & 0.144 & 0.390 & 0.387 & 0.392 & 0.394 & 0.395 & 0.396 & 0.402 \\
    \bottomrule
    \end{tabular}%
    }
    \caption{Accuracy before and after configuring our defense on ML-1M and Yelp.}
  \label{tab:accuracy_sample}%
\end{table*}%

\subsection{Recommendation Performance Evaluation} 
A good defense strategy cannot sacrifice recommendation performance by focusing on robustness. To this end, this section evaluates the recommendation performance of the defended model, as shown in Table \ref{tab:accuracy_sample}. Our defense method succeeds in sustaining the test accuracy of the base model. The test accuracy post-defense does not exhibit significant deterioration when compared to the pre-defense state. Results on the other two datasets are shown in Appendix E. 
We may understand the improvement of recommendation performance from two aspects: (1) The improved robustness implies cleanup of the influence of malicious data, which leads to better recommendation performance. (2) Vicinal training can make the model more focused on the current neighborhood and ensure the quality of learning.

\subsection{Inference Efficiency Evaluation}

\begin{table}
\centering
\begin{tabular}{c|c|c|c|c}
\midrule
Model & AutoRec & CDAE & MF & NSVD \\ \midrule
Origin (s) & 0.007 & 0.162 & 0.007 & 0.169 \\ \midrule
RVD (s) & 0.131 & 1.268 & 0.056 & 0.595 \\ \midrule
\end{tabular}
\caption{Inference time on Yelp per user.}
\label{tab:efficiency}
\end{table}

It is undeniable that the proposed defense method will reduce the inference time of the model. In this section, we compare the inference time required for each user. We take Yelp as an example since it is the largest dataset, as shown in Table \ref{tab:efficiency}. RVD is the total time in our method, and Origin is the time cost for pure prediction, i.e., the inference time of normal recommender systems. We can see our method costs more time than an ordinary recommender system. However, for one user, this delay is tolerable and negligible.

\begin{figure*}[!t]
	\centering
 \includegraphics[width=1 \textwidth]{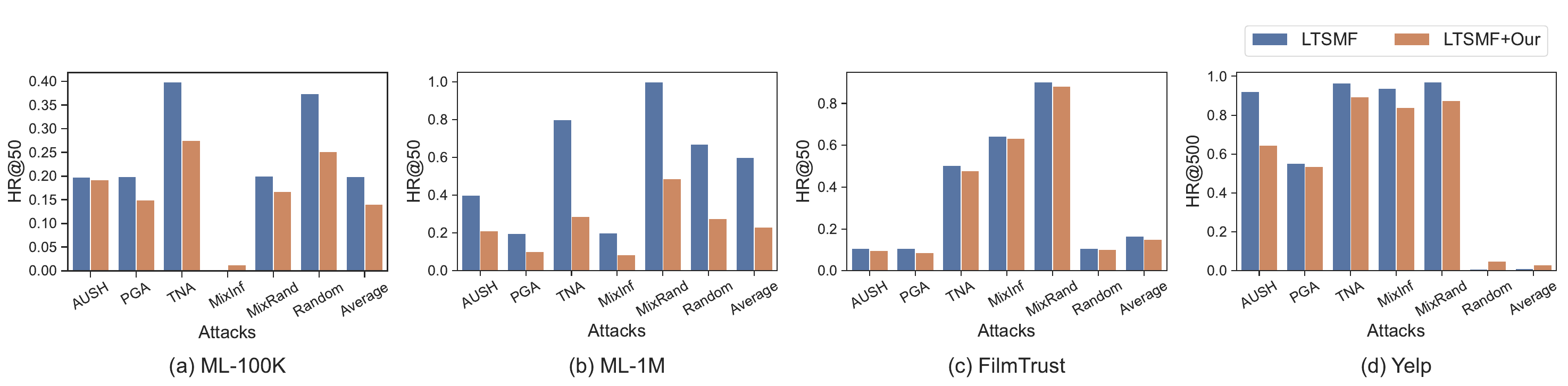}
	\caption{Target items' hit ratios on LTSMF before and after applying our method.}
		\label{fig:improve_ltsmf}
\end{figure*}
\begin{figure*}[!t]
	\centering
 \includegraphics[width=1 \textwidth]{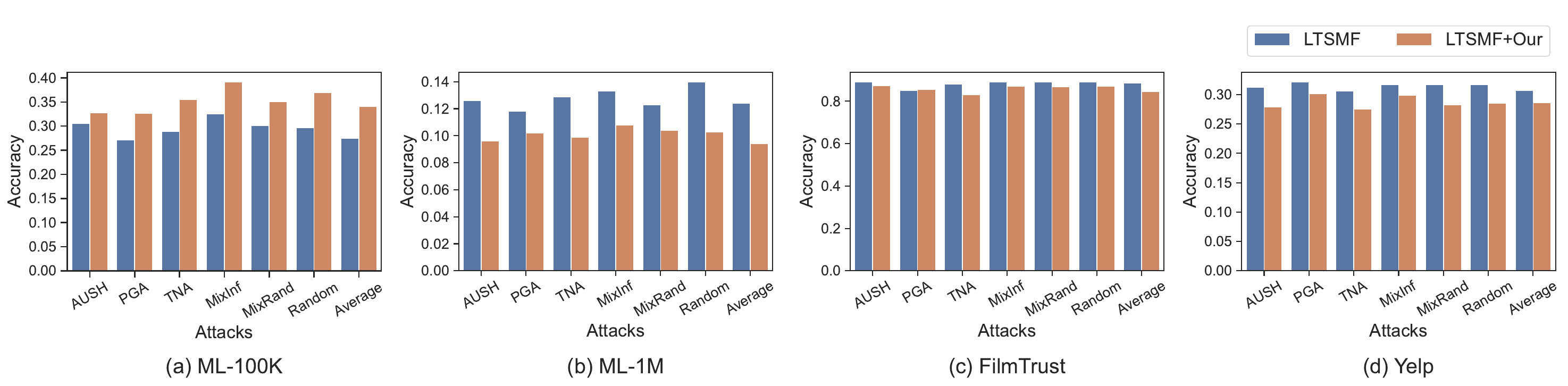}
	\caption{Accuracy of LTSMF before and after applying our method.}
		\label{fig:improve_ltsmf_accu}
\end{figure*}

\subsection{Enhancing Existing Defenses with Our Method}
Since our defense strategy focuses on the inference phase, this makes it easily configurable to any model-based or training defense strategy. This section demonstrates how our method can further bolster robustness when applied to some existing defenses.

Figure \ref{fig:improve_ltsmf} illustrates the robust performance when RVD is applied to LTSMF. Since the trimming mechanism of LTSMF impacts the convexity of the loss function, unlearning is rather challenging. Encouragingly, in most instances, the average hit ratio of target items after applying our method reduces, signifying an improvement in recommendation robustness. The result of applying our method on RMF is shown in Figure \ref{fig:improve_rmf} in Appendix E, which has similar findings.

Figure \ref{fig:improve_ltsmf_accu} further corroborates that our method enhances the robustness of the recommender system without a substantial sacrifice in accuracy. Despite the considerable improvement in defense against poisoning attacks, the recommender system maintains high test accuracy following the application of our method. This attests to the strength and efficiency of our approach, striking a desirable balance between system robustness and recommendation accuracy.

Upon aggregating all the results from our series of experiments, we confidently affirm that our proposed method significantly boosts the robustness of the recommender system, effectively defending against poisoning attacks. Remarkably, this considerable improvement in system security is not accompanied by a drastic decrease in accuracy.

\section{Conclusion}
In this study, we introduced a novel algorithm rooted in runtime fine-tuning, effectively bolsters robustness without incurring substantial accuracy costs. Moreover, we furnished a theoretical bound to gauge its capability to unlearn the influence of malicious users. Notably, the proposed method will not alter the model architecture or the training mode, which makes it more practical. Our experimental results showed that our method is more effective compared with baselines. Future research efforts will focus on exploring more efficient real-time defense strategies.

\bibliography{refs}

\begin{thebibliography}{35}
\providecommand{\natexlab}[1]{#1}

\bibitem[{Bampis et~al.(2017)Bampis, Rusu, Hajj, and Bovik}]{bampis2017robust}
Bampis, C.~G.; Rusu, C.; Hajj, H.; and Bovik, A.~C. 2017.
\newblock Robust matrix factorization for collaborative filtering in
  recommender systems.
\newblock In \emph{2017 51st Asilomar Conference on Signals, Systems, and
  Computers}.

\bibitem[{Baruch, Baruch, and
  Goldberg(2019)}]{DBLP:journals/corr/abs-1902-06156}
Baruch, M.; Baruch, G.; and Goldberg, Y. 2019.
\newblock A Little Is Enough: Circumventing Defenses For Distributed Learning.
\newblock In \emph{CoRR}.

\bibitem[{Borgnia et~al.(2021)Borgnia, Cherepanova, Fowl, Ghiasi, Geiping,
  Goldblum, Goldstein, and Gupta}]{borgnia2021strong}
Borgnia, E.; Cherepanova, V.; Fowl, L.; Ghiasi, A.; Geiping, J.; Goldblum, M.;
  Goldstein, T.; and Gupta, A. 2021.
\newblock Strong data augmentation sanitizes poisoning and backdoor attacks
  without an accuracy tradeoff.
\newblock In \emph{ICASSP 2021-2021 IEEE International Conference on Acoustics,
  Speech and Signal Processing (ICASSP)}.

\bibitem[{Cao and Gong(2017)}]{cao2017mitigating}
Cao, X.; and Gong, N.~Z. 2017.
\newblock Mitigating evasion attacks to deep neural networks via region-based
  classification.
\newblock In \emph{Proceedings of the 33rd Annual Computer Security
  Applications Conference}.

\bibitem[{Cheng and Hurley(2010)}]{cheng2010robust}
Cheng, Z.; and Hurley, N. 2010.
\newblock Robust collaborative recommendation by least trimmed squares matrix
  factorization.
\newblock In \emph{2010 22nd IEEE International Conference on Tools with
  Artificial Intelligence}.

\bibitem[{Fang et~al.(2019)Fang, Cao, Jia, and
  Gong}]{DBLP:journals/corr/abs-1911-11815}
Fang, M.; Cao, X.; Jia, J.; and Gong, N.~Z. 2019.
\newblock Local Model Poisoning Attacks to Byzantine-Robust Federated Learning.
\newblock In \emph{CoRR}.

\bibitem[{Fang, Gong, and Liu(2020)}]{fang2020influence}
Fang, M.; Gong, N.~Z.; and Liu, J. 2020.
\newblock Influence function based data poisoning attacks to top-n recommender
  systems.
\newblock In \emph{Proceedings of The Web Conference 2020}.

\bibitem[{Fang et~al.(2018)Fang, Yang, Gong, and Liu}]{fang2018poisoning}
Fang, M.; Yang, G.; Gong, N.~Z.; and Liu, J. 2018.
\newblock Poisoning attacks to graph-based recommender systems.
\newblock In \emph{Proceedings of the 34th annual computer security
  applications conference}.

\bibitem[{Fouss et~al.(2007)Fouss, Pirotte, Renders, and Saerens}]{4072747}
Fouss, F.; Pirotte, A.; Renders, J.-m.; and Saerens, M. 2007.
\newblock Random-Walk Computation of Similarities between Nodes of a Graph with
  Application to Collaborative Recommendation.
\newblock In \emph{IEEE Transactions on Knowledge and Data Engineering}.

\bibitem[{Golbeck and Hendler(2006)}]{1593032}
Golbeck, J.; and Hendler, J. 2006.
\newblock FilmTrust: movie recommendations using trust in web-based social
  networks.
\newblock In \emph{CCNC 2006. 2006 3rd IEEE Consumer Communications and
  Networking Conference, 2006.}

\bibitem[{Goodfellow et~al.(2014)Goodfellow, Pouget-Abadie, Mirza, Xu,
  Warde-Farley, Ozair, Courville, and Bengio}]{goodfellow2014generative}
Goodfellow, I.~J.; Pouget-Abadie, J.; Mirza, M.; Xu, B.; Warde-Farley, D.;
  Ozair, S.; Courville, A.; and Bengio, Y. 2014.
\newblock Generative Adversarial Networks.

\bibitem[{GroupLens(1998)}]{movielens100k}
GroupLens. 1998.
\newblock MovieLens 100K Dataset.

\bibitem[{GroupLens(2003)}]{movielens1m}
GroupLens. 2003.
\newblock MovieLens 1M Dataset.

\bibitem[{He et~al.(2020)He, Deng, Wang, Li, Zhang, and
  Wang}]{10.1145/3397271.3401063}
He, X.; Deng, K.; Wang, X.; Li, Y.; Zhang, Y.; and Wang, M. 2020.
\newblock LightGCN: Simplifying and Powering Graph Convolution Network for
  Recommendation.

\bibitem[{He et~al.(2018)He, He, Du, and Chua}]{he2018adversarial}
He, X.; He, Z.; Du, X.; and Chua, T.-S. 2018.
\newblock Adversarial personalized ranking for recommendation.
\newblock In \emph{The 41st International ACM SIGIR conference on research \&
  development in information retrieval}.

\bibitem[{He, Zhang, and Lee(2019)}]{10.1145/3359789.3359824}
He, Z.; Zhang, T.; and Lee, R.~B. 2019.
\newblock Model Inversion Attacks against Collaborative Inference.
\newblock In \emph{Proceedings of the 35th Annual Computer Security
  Applications Conference}.

\bibitem[{Hidano et~al.(2017)Hidano, Murakami, Katsumata, Kiyomoto, and
  Hanaoka}]{hidano2017model}
Hidano, S.; Murakami, T.; Katsumata, S.; Kiyomoto, S.; and Hanaoka, G. 2017.
\newblock Model inversion attacks for prediction systems: Without knowledge of
  non-sensitive attributes.
\newblock In \emph{2017 15th Annual Conference on Privacy, Security and Trust
  (PST)}.

\bibitem[{Koren, Bell, and Volinsky(2009)}]{5197422}
Koren, Y.; Bell, R.; and Volinsky, C. 2009.
\newblock In \emph{Matrix Factorization Techniques for Recommender Systems}.

\bibitem[{Lam and Riedl(2004)}]{lam2004shilling}
Lam, S.~K.; and Riedl, J. 2004.
\newblock Shilling recommender systems for fun and profit.
\newblock In \emph{Proceedings of the 13th international conference on World
  Wide Web}.

\bibitem[{Li et~al.(2016)Li, Wang, Singh, and Vorobeychik}]{li2016data}
Li, B.; Wang, Y.; Singh, A.; and Vorobeychik, Y. 2016.
\newblock Data poisoning attacks on factorization-based collaborative
  filtering.
\newblock In \emph{Advances in neural information processing systems}.

\bibitem[{Lin et~al.(2020)Lin, Chen, Li, Xiao, Li, and Yang}]{lin2020attacking}
Lin, C.; Chen, S.; Li, H.; Xiao, Y.; Li, L.; and Yang, Q. 2020.
\newblock Attacking recommender systems with augmented user profiles.
\newblock In \emph{Proceedings of the 29th ACM international conference on
  information \& knowledge management}.

\bibitem[{Pang et~al.(2022)Pang, Wu, Shen, Zhang, Wei, Xu, Chang, Long, and
  Pei}]{Pang_2022}
Pang, Y.; Wu, L.; Shen, Q.; Zhang, Y.; Wei, Z.; Xu, F.; Chang, E.; Long, B.;
  and Pei, J. 2022.
\newblock Heterogeneous Global Graph Neural Networks for Personalized
  Session-based Recommendation.
\newblock In \emph{Proceedings of the Fifteenth {ACM} International Conference
  on Web Search and Data Mining}.

\bibitem[{Paterek(2007)}]{paterek2007improving}
Paterek, A. 2007.
\newblock Improving regularized singular value decomposition for collaborative
  filtering.
\newblock In \emph{Proceedings of KDD cup and workshop}.

\bibitem[{Peri et~al.(2020)Peri, Gupta, Huang, Fowl, Zhu, Feizi, Goldstein, and
  Dickerson}]{peri2020deep}
Peri, N.; Gupta, N.; Huang, W.~R.; Fowl, L.; Zhu, C.; Feizi, S.; Goldstein, T.;
  and Dickerson, J.~P. 2020.
\newblock Deep k-nn defense against clean-label data poisoning attacks.
\newblock In \emph{Computer Vision--ECCV 2020 Workshops: Glasgow, UK, August
  23--28, 2020, Proceedings, Part I 16}.

\bibitem[{Popkov(2005)}]{unconstrained2005}
Popkov, A.~Y. 2005.
\newblock Gradient Methods for Nonstationary Unconstrained Optimization
  Problems.

\bibitem[{Sarwar et~al.(2001)Sarwar, Karypis, Konstan, and
  Riedl}]{10.1145/371920.372071}
Sarwar, B.; Karypis, G.; Konstan, J.; and Riedl, J. 2001.
\newblock Item-Based Collaborative Filtering Recommendation Algorithms.
\newblock In \emph{Proceedings of the 10th International Conference on World
  Wide Web}.

\bibitem[{Sedhain et~al.(2015)Sedhain, Menon, Sanner, and
  Xie}]{sedhain2015autorec}
Sedhain, S.; Menon, A.~K.; Sanner, S.; and Xie, L. 2015.
\newblock Autorec: Autoencoders meet collaborative filtering.
\newblock In \emph{Proceedings of the 24th international conference on World
  Wide Web}.

\bibitem[{Steinhardt, Koh, and Liang(2017)}]{steinhardt2017certified}
Steinhardt, J.; Koh, P. W.~W.; and Liang, P.~S. 2017.
\newblock Certified defenses for data poisoning attacks.
\newblock In \emph{Advances in neural information processing systems}.

\bibitem[{van~der Maaten and Hinton(2008)}]{JMLR:v9:vandermaaten08a}
van~der Maaten, L.; and Hinton, G. 2008.
\newblock Visualizing Data using t-SNE.
\newblock \emph{Journal of Machine Learning Research}, 9(86): 2579--2605.

\bibitem[{Wang et~al.(2022)Wang, Ma, Wang, Hu, Qin, and Ren}]{wang2022threats}
Wang, Z.; Ma, J.; Wang, X.; Hu, J.; Qin, Z.; and Ren, K. 2022.
\newblock Threats to Training: A Survey of Poisoning Attacks and Defenses on
  Machine Learning Systems.
\newblock In \emph{ACM Computing Surveys}.

\bibitem[{Wu et~al.(2023)Wu, Lian, Ge, Zhu, and Chen}]{10122715}
Wu, C.; Lian, D.; Ge, Y.; Zhu, Z.; and Chen, E. 2023.
\newblock Influence-Driven Data Poisoning for Robust Recommender Systems.
\newblock In \emph{IEEE Transactions on Pattern Analysis and Machine
  Intelligence}.

\bibitem[{Wu et~al.(2016)Wu, DuBois, Zheng, and Ester}]{wu2016collaborative}
Wu, Y.; DuBois, C.; Zheng, A.~X.; and Ester, M. 2016.
\newblock Collaborative denoising auto-encoders for top-n recommender systems.
\newblock In \emph{Proceedings of the ninth ACM international conference on web
  search and data mining}.

\bibitem[{Yelp(2004)}]{yelp}
Yelp. 2004.
\newblock Yelp Dataset.

\bibitem[{Zhang et~al.(2017)Zhang, Cisse, Dauphin, and
  Lopez-Paz}]{zhang2017mixup}
Zhang, H.; Cisse, M.; Dauphin, Y.~N.; and Lopez-Paz, D. 2017.
\newblock mixup: Beyond empirical risk minimization.
\newblock In \emph{arXiv preprint arXiv:1710.09412}.

\bibitem[{Zhang et~al.(2020)Zhang, Lou, Chen, Yuan, Li, Johnsten, and
  Tzeng}]{10.1007/978-3-030-58951-6_23}
Zhang, Y.; Lou, J.; Chen, L.; Yuan, X.; Li, J.; Johnsten, T.; and Tzeng, N.-F.
  2020.
\newblock Towards Poisoning the Neural Collaborative Filtering-Based
  Recommender Systems.
\newblock In \emph{Computer Security -- ESORICS 2020}.

\end{thebibliography}

\clearpage
\appendix
\section{A. Proof of Theorems}
Before the proof, we first introduce two necessary conclusions to assist the proof.

Lemma \ref{lemma1} and Theorem \ref{theorem1} are proposed in~\cite{unconstrained2005}.
\begin{lemma}[~\cite{unconstrained2005}]
\label{lemma1}
Suppose $\mathcal{D}_0$ is a dataset of size $n$, $\mathcal{D}_1$ is a new dataset by adding or deleting an entry in $\mathcal{D}_0$. $f:\Theta\to\mathbb{R}$ is the loss function for parameter $\theta\in\Theta$. $f_{\mathcal{D}}$ is that loss function on dataset $\mathcal{D}$. Both $f_{\mathcal{D}_0}$ and $f_{\mathcal{D}_1}$ are m-convex and L-Lipschitz. Let $\theta_{\mathcal{D}_i}^{*}=\text{argmin}_{\theta\in\Theta}f_{\mathcal{D}_i}(\theta)$, then we have
$$
\|\theta_{\mathcal{D}_0}^{*}-\theta_{\mathcal{D}_1}^{*}\|_2\le\frac{2L}{mn}.
$$
\end{lemma}

\begin{theorem}[~\cite{unconstrained2005}]
\label{theorem1}
Let $f$ be m-strongly convex and M-smooth, $\theta^{*}=\text{argmin}_{\theta\in\Theta}f(\theta)$. After $T$ steps of gradient descent with step size $\eta=\frac{2}{m+M}$, 
$$
\|\theta_T-\theta^{*}\|_2\le\left(\frac{M-m}{M+m}\right)^T\|\theta_0-\theta^{*}\|_2
$$
\end{theorem}

\subsection{Proof of Theorem \ref{theorem2}}
\begin{proof}
According to Lemma \ref{lemma1} and the triangular inequality, we have
$$
\|\theta_{\mathcal{D}_0}^{*}-\theta_{\mathcal{D}_1}^{*}\|_2\le\frac{2L}{mn}+\frac{2L}{m(n-1)}+\cdots+\frac{2L}{m(k+1)}.
$$
When n is large, we have
$$
\|\theta_{\mathcal{D}_0}^{*}-\theta_{\mathcal{D}_1}^{*}\|_2=O\left(\frac{2L}{m}\log\left(\frac{n}{k}\right)\right).
$$
Suppose the model parameter after implementing the first $T$ steps of gradient descent is $\hat{\theta}_{\mathcal{D}_0}$. According to Theorem \ref{theorem1}, we have
$$
\|\hat{\theta}_{\mathcal{D}_0}-\theta_{\mathcal{D}_0}^{*}\|_2\le\gamma^T\|\theta_{\mathcal{D}_0}-\theta_{\mathcal{D}_0}^{*}\|_2\le\frac{2L\gamma^I\|\theta_{\mathcal{D}_0}-\theta_{\mathcal{D}_0}^{*}\|_2}{Dmn}\le\frac{2L\gamma^I}{mn}.
$$
Therefore,
\begin{align*}
    \|\hat{\theta}_{\mathcal{D}_1}-\theta_{\mathcal{D}_1}^{*}\|_2&\le\gamma^I\|\hat{\theta}_{\mathcal{D}_0}-\theta_{\mathcal{D}_1}^{*}\|_2 \\
    &\le\gamma^I(\|\hat{\theta}_{\mathcal{D}_0}-\theta_{\mathcal{D}_0}^{*}\|_2+\|\theta_{\mathcal{D}_0}^{*}-\theta_{\mathcal{D}_1}^{*}\|_2) \\
    &\le\frac{2L\gamma^I}{m}\left(\frac{1}{n}+\frac{1}{n}+\frac{1}{n-1}+\cdots+\frac{1}{k+1}\right) \\
    &=O\left(\frac{2L\gamma^I}{m}\log\left(\frac{n}{k}\right)\right).
\end{align*}
Since $f$ is M-smooth and strongly convex, 
\begin{align*}
    |f_{\mathcal{D}_1}(\hat{\theta}_{\mathcal{D}_1})-f_{\mathcal{D}_1}(\theta_{\mathcal{D}_1}^{*})|&\le\frac{M}{2}\|\hat{\theta}_{\mathcal{D}_1}-\theta_{\mathcal{D}_1}^{*}\|_2^2\\
    &=O\left(\frac{L^2\gamma^{2I}}{m^2M}\log^2\left(\frac{n}{k}\right)\right).
\end{align*}
\end{proof}
\subsection{Proof of Theorem \ref{theorem3}}
\begin{proof}
Denote $g(\theta)=f(\theta)+\frac{m}{2}\|\theta\|_2^2$, then $g$ is m-strongly-convex, (L+mD)-Lipschitz and (M+m)-smooth.
    Denote $\theta_{\mathcal{D}_i}^{*r}=\text{argmin}_{\theta\in\Theta}g_{\mathcal{D}_i}(\theta)$. According to Theorem \ref{theorem2}, we have
    $$
    \|\hat{\theta}_{\mathcal{D}_1}-\theta_{\mathcal{D}_1}^{*r}\|_2=O\left(\frac{(L+mD)\gamma^I}{m}\log\left(\frac{n}{k}\right)\right).
    $$
    Therefore, 
    \begin{flalign*}
        &f_{\mathcal{D}_1}(\hat{\theta}_{\mathcal{D}_1})-f_{\mathcal{D}_1}(\theta_{\mathcal{D}_1}^{*})=f_{\mathcal{D}_1}(\hat{\theta}_{\mathcal{D}_1})-f_{\mathcal{D}_1}(\theta_{\mathcal{D}_1}^{*r})+f_{\mathcal{D}_1}(\theta_{\mathcal{D}_1}^{*r}) \\
        &-f_{\mathcal{D}_1}(\theta_{\mathcal{D}_1}^{*}) \\
        &\le\nabla f_{\mathcal{D}_1}(\theta_{\mathcal{D}_1}^{*r})^T(\hat{\theta}_{\mathcal{D}_1}-\theta_{\mathcal{D}_1}^{*r})+\frac{M}{2}\|\hat{\theta}_{\mathcal{D}_1}-\theta_{\mathcal{D}_1}^{*r}\|_2^2+f_{\mathcal{D}_1}(\theta_{\mathcal{D}_1}^{*r}) \\
        &-f_{\mathcal{D}_1}(\theta_{\mathcal{D}_1}^{*}) \\
        &=\frac{M}{2}\|\hat{\theta}_{\mathcal{D}_1}-\theta_{\mathcal{D}_1}^{*r}\|_2^2+m\theta_{\mathcal{D}_1}^{*rT}(\theta_{\mathcal{D}_1}^{*r}-\hat{\theta}_{\mathcal{D}_1})+f_{\mathcal{D}_1}(\theta_{\mathcal{D}_1}^{*r}) \\
        &-f_{\mathcal{D}_1}(\theta_{\mathcal{D}_1}^{*}) \\
        &\le\frac{M}{2}\|\hat{\theta}_{\mathcal{D}_1}-\theta_{\mathcal{D}_1}^{*r}\|_2^2+mD^2+f_{\mathcal{D}_1}(\theta_{\mathcal{D}_1}^{*r})-f_{\mathcal{D}_1}(\theta_{\mathcal{D}_1}^{*}) \\
        &=\frac{M}{2}\|\hat{\theta}_{\mathcal{D}_1}-\theta_{\mathcal{D}_1}^{*r}\|_2^2+mD^2+g_{\mathcal{D}_1}(\theta_{\mathcal{D}_1}^{*r})-\frac{m}{2}\|\theta_{\mathcal{D}_1}^{*r}\|_2^2-f_{\mathcal{D}_1}(\theta_{\mathcal{D}_1}^{*}) \\
        &\le\frac{M}{2}\|\hat{\theta}_{\mathcal{D}_1}-\theta_{\mathcal{D}_1}^{*r}\|_2^2+mD^2+g_{\mathcal{D}_1}(\theta_{\mathcal{D}_1}^{*r})-\frac{m}{2}\|\theta_{\mathcal{D}_1}^{*}\|_2^2-f_{\mathcal{D}_1}(\theta_{\mathcal{D}_1}^{*}) \\
        &=\frac{M}{2}\|\hat{\theta}_{\mathcal{D}_1}-\theta_{\mathcal{D}_1}^{*r}\|_2^2+mD^2+\frac{m}{2}(\|\theta_{\mathcal{D}_1}^*\|_2^2-\|\theta_{\mathcal{D}_1}\|_2^2) \\
        &=O\left(\frac{(L+mD)^2\gamma^{2I}}{m^2(M+m)}\log^2\left(\frac{n}{k}\right)+mD^2\right)
    \end{flalign*}
    There are three inequalities in the above formula. The first one follows from $f$ is M-smooth. The second one follows from the fact that $\theta_{\mathcal{D}_1}^{*rT}(\theta_{\mathcal{D}_1}^{*r}-\hat{\theta}_{\mathcal{D}_1})\le D^2$, and the third inequality follows from the minimal property of $\theta_{\mathcal{D}_1}^{*r}$. Equalities in the above formula are easy to derive.
\end{proof}

\begin{table*}[htbp]
  \centering
  \footnotesize
  \renewcommand\arraystretch{0.85}
  \setlength{\tabcolsep}{0.006\linewidth}{
    \begin{tabular}{c|c|c|c|c|c|c|c|c|c|c|c|c|c|c}
    \toprule
    Dataset & \multicolumn{7}{c|}{ML-100K (HR@50)}                          & \multicolumn{7}{c}{FilmTrust (HR@50)} \\
    \midrule
    Attack & AUSH  & PGA   & TNA   & MixInf & MixRand & Random & Average & AUSH  & PGA   & TNA   & MixInf & MixRand & Random & Average \\
    \midrule
    no defense & 0.994 & 0.994 & 0.995 & 0.998 & 0.997 & 0.950 & 0.996 & 0.864 & 0.542 & 0.869 & 0.886 & 0.900 & 0.690 & 0.565 \\
    APR   & 0.852 & 0.787 & 0.806 & 0.973 & 0.885 & 0.348 & 0.775 & 0.664 & 0.184 & 0.676 & 0.792 & 0.794 & 0.256 & 0.199 \\
    LTSMF & 0.198 & 0.199 & 0.399 & 0.000 & 0.200 & 0.374 & 0.199 & \textbf{0.108} & \textbf{0.107} & \textbf{0.504} & \textbf{0.644} & 0.903 & \textbf{0.107} & 0.166 \\
    RMF   & \textbf{0.025} & \textbf{0.025} & 0.026 & \textbf{0.025} & \textbf{0.016} & 0.012 & 0.022 & 0.865 & 0.859 & 0.737 & 0.715 & \textbf{0.727} & 0.864 & 0.851 \\
    Our   & 0.069 & 0.032 & \textbf{0.012} & 0.294 & 0.202 & \textbf{0.011} & \textbf{0.018} & 0.362 & 0.128 & 0.849 & 0.873 & 0.891 & 0.108 & \textbf{0.136} \\
    \bottomrule
    \end{tabular}%
    }
    \caption{Target items' hit ratios after different defenses on ML-100K and FilmTrust. Notably, their hit ratios are 0 when there is no attack. Therefore, the lower the hit ratio, the better the defense performance.}
  \label{tab:compare}%
\end{table*}%

\begin{table*}[htbp]
  \centering
  \footnotesize
  \renewcommand\arraystretch{0.85}
  \setlength{\tabcolsep}{0.0052\linewidth}{
    \begin{tabular}{c|c|c|c|c|c|c|c|c|c|c|c|c|c|c|c}
    \toprule
    \multicolumn{2}{c|}{Dataset} & \multicolumn{7}{c|}{ML-100K (HR@50)}                          & \multicolumn{7}{c}{FilmTrust (HR@50)} \\
    \midrule
    Model & Attack & AUSH  & PGA   & TNA   & MixInf & MixRand & Random & Average & AUSH  & PGA   & TNA   & MixInf & MixRand & Random & Average \\
    \midrule
    \multirow{2}[2]{*}{AutoRec} & before & 0.979 & 0.996 & 0.994 & 0.597 & 0.992 & 0.997 & 0.674 & 0.849 & 0.314 & 0.404 & 0.808 & 0.365 & 0.649 & 0.593 \\
          & after & 0.000 & 0.001 & 0.000 & 0.137 & 0.117 & 0.000 & 0.000 & 0.073 & 0.072 & 0.073 & 0.117 & 0.089 & 0.067 & 0.075 \\
    \midrule
    \multirow{2}[2]{*}{CDAE} & before & 0.601 & 0.619 & 0.602 & 0.599 & 0.600 & 0.801 & 0.844 & 0.299 & 0.614 & 0.789 & 0.624 & 0.788 & 0.282 & 0.312 \\
          & after & 0.018 & 0.020 & 0.007 & 0.195 & 0.145 & 0.003 & 0.022 & 0.122 & 0.097 & 0.108 & 0.232 & 0.206 & 0.086 & 0.104 \\
    \midrule
    \multirow{2}[2]{*}{NSVD} & before & 0.925 & 0.742 & 0.913 & 0.976 & 0.985 & 0.803 & 0.905 & 0.107 & 0.106 & 0.109 & 0.839 & 0.831 & 0.115 & 0.109 \\
          & after & 0.013 & 0.006 & 0.015 & 0.050 & 0.053 & 0.010 & 0.009 & 0.076 & 0.063 & 0.060 & 0.071 & 0.071 & 0.075 & 0.053 \\
    \bottomrule
    \end{tabular}%
    }
    \caption{Target items' hit ratios under different recommendation models on ML-100K and FilmTrust. Their hit ratios are 0 when there's no attack.}
  \label{tab:hr}%
\end{table*}%

\section{B. Details of Poisoning Attacks} \label{sec:appB}
This section provides detailed illustrations of poisoning attacks used in our experiment.

1. \textbf{AUSH}~\cite{lin2020attacking}: This type of attack utilizes the concept of Generative Adversarial Networks (GANs)~\cite{goodfellow2014generative}. The key aspect of this approach lies in its capacity to generate inconspicuous or 'imperceptible' users. These synthetic users are then employed in assigning the highest possible rating to a designated target item.

2. \textbf{PGA}~\cite{li2016data}: In PGA attack, the attacker has a distinct objective function that encapsulates the task of poisoning. Stochastic Gradient Descent (SGD) is then employed to minimize this function. The method focuses on top $k$ items—determined based on their high ratings—and designates them as filler items. This attack manipulates the system subtly by altering the ratings of the chosen filler items.

3. \textbf{TNA}~\cite{fang2020influence}: The TNA attack takes an approach focused on the users. It identifies the top $k$ influential users within the dataset. These influential users are then leveraged to assign maximum possible ratings to the targeted items. Once assigned, a series of fine-tunings are made to these user profiles, subtly increasing the ratings of the targeted items, thereby ensuring their inclusion in the top-rated items.

4. \textbf{MixInf}~\cite{10122715}: This attack includes two key components: an influence-based threat estimator and a generator, Usermix. The influence-based threat estimator can evaluate the detriment of poisoning users to the system without retraining the model. Usermix is a model-agnostic method for generating reasonable users by fusing users' interests through linear interpolating two rating vectors. The proposed Infmix integrates them to create malicious but non-notable users for poisoning.

5. \textbf{MixRand}~\cite{10122715}: This attack is a non-influential version of Infmix that uses Usermix to generate users and assigns the maximum rating directly to target items.

6. \textbf{RANDOM}~\cite{lam2004shilling}: In this method, the attacker assigns the maximum possible rating to the target items. Subsequently, a random selection of filler items is made. These items are then rated according to a distribution that is designed to mimic the rating distribution in the overall dataset.

7. \textbf{AVERAGE}~\cite{lam2004shilling}: The AVERAGE attack shares a similar initial approach with the RANDOM attack; it assigns maximum ratings to target items and randomly selects filler items. However, instead of following a pre-determined distribution, this method assigns completely random ratings to these filler items.
\section{C. Details of Baseline Defense} \label{sec:appC}
This part gives technical details of defense strategies used in our experiment.

{\bfseries Adversarial Personalized Ranking (APR)}~\cite{he2018adversarial}: This strategy optimizes the model by playing a minimax game, defined as follows: 
$$
L_{\text{APR}}(D|\Theta) = L_{\text{BPR}}(D|\Theta) + \lambda L_{\text{BPR}}(D|\Theta + \Delta_{\text{adv}}),
$$ 
where $\Delta_{\text{adv}} = \arg \max_{\Delta,||\Delta||\le\epsilon}L_{\text{BPR}}(D|\hat{\Theta}+\Delta)$, and $L_{\text{BPR}}$ is defined as $L_{\text{BPR}} = \sum_{(u,i,d)\in D}-\ln\sigma(\hat{y}_{ui}(\Theta)-\hat{y}_{uj}(\Theta))+\lambda_{\Theta}||\Theta||^2$. In each training round, the model first trains the parameter $\Theta$ to fit the model, and then trains the adversarial perturbation $\Delta_{\text{adv}}$ to increase the loss.

{\bfseries Least Trimmed Square Matrix Factorization (LTSMF)}~\cite{cheng2010robust}: This approach uses a robust loss function to enhance robustness. For matrix factorization, the loss function is $\sum_{r_{u,i}>0}(r_{u,i}-\mathbf{q}_i^T\mathbf{p}_u)^2+\lambda*reg$, where $r_{u,i}$ is the rating of user u with item i, and $\mathbf{p_u}$ and $\mathbf{q}_i$ are the feature vectors of user u and item i, respectively. This method trims the first term. It disregards $(u,i)$ pairs for which $(r_{u,i}-\mathbf{q}_i^T\mathbf{p}_u)^2$ falls within the top largest $k$ percent of all $(u,i)$ pairs.

{\bfseries Robust Matrix Factorization (RMF)}~\cite{bampis2017robust}: This method employs a different optimization target:
$$
U=\arg\min_U\|Y-A(UU^TY)\|_F^2,
$$
where $U$ is the model parameter, $Y$ is the entire user-item rating matrix, $A$ is an operator that only preserves informative entries (other non-rated entries are reset to 0), and $\|\cdot\|_F$ represents the Frobenius norm.
\section{D. Details of Evaluation Metrics} \label{sec:appD}
Reading the literal description of our evaluation metrics may be confusing. Therefore, in this section, we provide formulas describing how these metrics are exactly computed.

\textbf{Accuracy:} For each user $u \in U$, where $U$ is the set of all users, denote its test item as $i_u$. The accuracy can be given by:

\[
\text{Accuracy} = \frac{1}{|U|}\sum_{u\in U}\mathbb{I}\left[i_u\in R_u\right],
\]

where $R_u$ represents the set of recommended items to user $u$ from $I_u^c$.

\textbf{Hit@k (HR@k):} The hit@k metric quantifies how frequently the targeted items in a poisoning attack are recommended among the top-k recommendations. For each target item $i \in T$, where $T$ is the set of targeted items, we compute the hit ratio (HR) as the proportion of users to whom the item $i$ is recommended. Only those users for whom $i$ is a non-interacted item are taken into account, represented as $U_i$. Formally, the hit ratio for the target item $i$ is given by:

$$
\text{HR}_i = \frac{1}{|U_i|}\sum_{u\in U_i}\mathbb{I}\left[i\in R_u\right].
$$

The overall hit@k is then the average HR over all target items.
\section{E. Additional Experimental Results}

\begin{table*}[htbp]
  \centering
  \footnotesize
  \renewcommand\arraystretch{0.85}
  \setlength{\tabcolsep}{0.0052\linewidth}{
    \begin{tabular}{c|c|c|c|c|c|c|c|c|c|c|c|c|c|c|c}
    \toprule
    \multicolumn{2}{c|}{Dataset} & \multicolumn{7}{c|}{ML-100K}                            & \multicolumn{7}{c}{FilmTrust} \\
    \midrule
    Model & Attack & AUSH  & PGA   & TNA   & MixInf & MixRand & Random & Average & AUSH  & PGA   & TNA   & MixInf & MixRand & Random & Average \\
    \midrule
    \multirow{2}[2]{*}{AutoRec} & before & 0.224 & 0.206 & 0.201 & 0.228 & 0.212 & 0.229 & 0.206 & 0.846 & 0.854 & 0.826 & 0.835 & 0.835 & 0.852 & 0.861 \\
          & after & 0.343 & 0.307 & 0.316 & 0.328 & 0.314 & 0.290 & 0.333 & 0.854 & 0.857 & 0.854 & 0.857 & 0.844 & 0.846 & 0.846 \\
    \midrule
    \multirow{2}[2]{*}{CDAE} & before & 0.171 & 0.159 & 0.183 & 0.165 & 0.170 & 0.157 & 0.158 & 0.854 & 0.857 & 0.834 & 0.868 & 0.861 & 0.864 & 0.868 \\
          & after & 0.219 & 0.201 & 0.195 & 0.211 & 0.244 & 0.228 & 0.208 & 0.850 & 0.848 & 0.830 & 0.854 & 0.850 & 0.832 & 0.854 \\
    \midrule
    \multirow{2}[2]{*}{MF} & before & 0.162 & 0.148 & 0.170 & 0.198 & 0.170 & 0.185 & 0.179 & 0.866 & 0.810 & 0.864 & 0.841 & 0.868 & 0.870 & 0.848 \\
          & after & 0.204 & 0.217 & 0.231 & 0.204 & 0.215 & 0.186 & 0.224 & 0.828 & 0.816 & 0.839 & 0.805 & 0.854 & 0.848 & 0.837 \\
    \midrule
    \multirow{2}[2]{*}{NSVD} & before & 0.292 & 0.334 & 0.294 & 0.371 & 0.343 & 0.298 & 0.284 & 0.866 & 0.864 & 0.870 & 0.859 & 0.861 & 0.864 & 0.866 \\
          & after & 0.401 & 0.380 & 0.393 & 0.379 & 0.389 & 0.405 & 0.374 & 0.834 & 0.830 & 0.830 & 0.832 & 0.839 & 0.835 & 0.830 \\
    \bottomrule
    \end{tabular}%
    }
    \caption{Accuracy before and after configuring our defense on ML-100K and FilmTrust.}
  \label{tab:accuracy}%
\end{table*}%

\begin{figure*}[htbp]
	\centering
        \includegraphics[width=1 \textwidth]{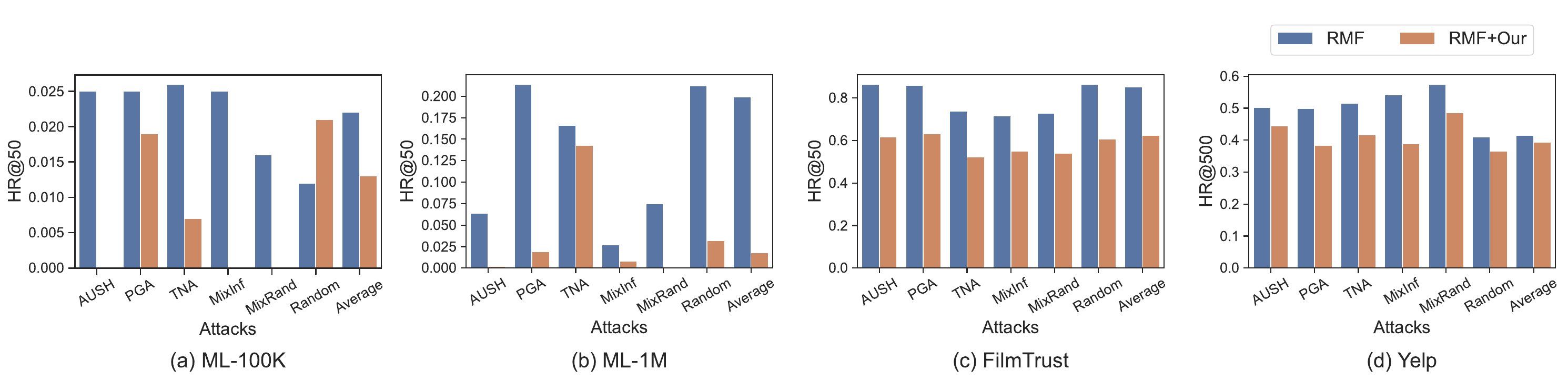}
	\caption{Target items' hit ratios on RMF before and after applying our method.}
		\label{fig:improve_rmf}
\end{figure*}

\begin{figure*}[htbp]
	\centering
 \includegraphics[width=1 \textwidth]{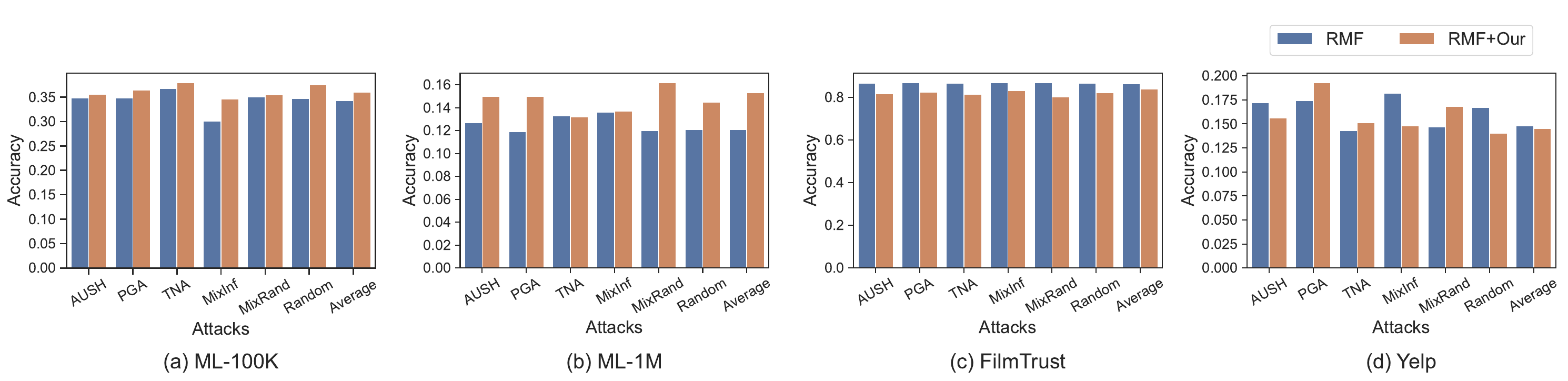}
	\caption{Accuracy of RMF before and after applying our method.}
		\label{fig:improve_rmf_accu}
\end{figure*}

\begin{figure*}[htbp]
    \centering
    \includegraphics[width = 0.6 \textwidth]{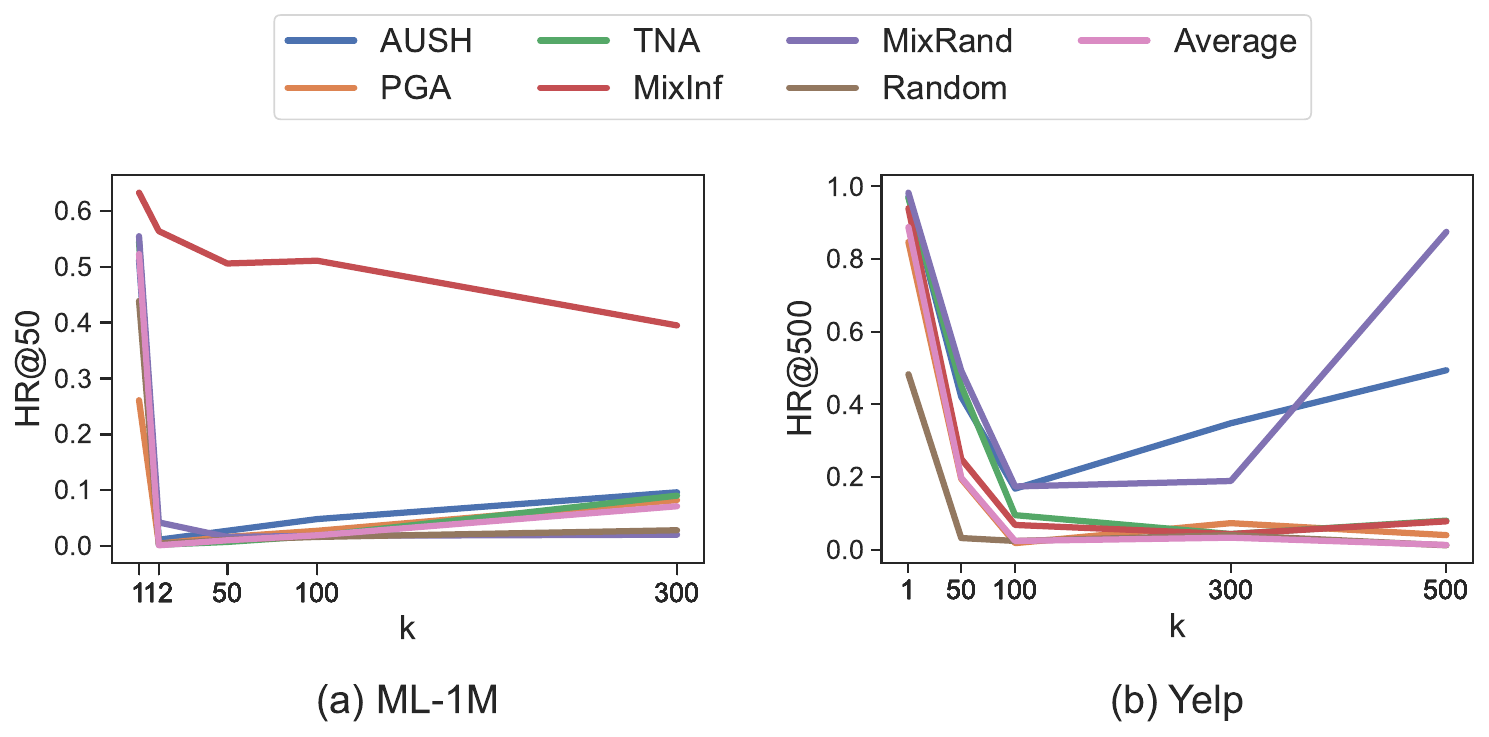}
    \caption{Comparison of defensive performance (hit ratio) under different $k$.}
    \label{fig:ablation}
\end{figure*}

\noindent\textbf{Additional results of defense performance comparison}. Table \ref{tab:compare} shows our evaluation of the target hit ratio on ML-100K and FilmTrust after different defenses. The base model is still MF. We can see that our method ranks top 2 in most cases while ranking first in over half of all cases, significantly surpassing APR. For other methods which are comparative to ours in some cases, we later demonstrate that by applying our method on them, their robustness can further be improved.

\noindent\textbf{Additional results of defense performance in other recommendation models}. Table \ref{tab:hr} shows the target hit ratio before and after our defense on the other three recommendation models (AutoRec, CDAE and NSVD) on ML-100M and FilmTrust. From the table, we can conclude that in most cases, our method significantly lowers the target hit ratio. For the sparse dataset, ML-100K, this is particularly obvious--in some cases, the target hit ratio directly drops from 0.99 to 0. This is because in such datasets, the embedding space is sparse and neighbors of a user are more likely to be benign users--malicious users reside in another region in the sparse space that is distant from benign users.

\noindent\textbf{Additional results of recommendation performance.} Table \ref{tab:accuracy} shows the recommendation performance before and after our defense on ML-100K and FilmTrust. For ML-100K, the test accuracy improves because malicious users are filtered too thoroughly. Just like what is discussed in the experiment part, the test accuracy may increase in such cases. For FilmTrust, it is more challenging to defend. Therefore, the test accuracy decreases, but on an acceptable scale.

\noindent\textbf{Results of applying our method on RMF}. We also try to apply our method on RMF to see whether \alg{} further improves the robustness of RMF. Figure \ref{fig:improve_rmf} shows the result. According to the figure, our method improves RMF in all cases. For instance when ML-100K is poisoned by AUSH, RMF lowers the target hit ratio to 0.025. However after applying our method on RMF, the target hit ratio drops to 0. Figure \ref{fig:improve_ltsmf_accu} shows the accuracy before and after applying our method. Still, the accuracy does not drop too much after using our method.

\noindent\textbf{Ablation study: impacts on different $k$}.
Our method has a parameter $k$ indicating how many neighbors are selected for each user to fine-tune the model. From the theoretical part, we concluded that $k$ should be large enough to ensure the unlearning ability. However, $k$ cannot be too large because there will be more malicious users in those $k$ neighbors. In this part, we choose different $k$ and explore the sensitivity of our result depending on $k$.

Figure \ref{fig:ablation} demonstrates our result. We can see when $k$ is too large or too small, the defense effect is not optimal. However in most cases, when $k$ is in a reasonable interval (not extremely large or extremely small), the defense effect is relatively stable even if $k$ is not optimal. This result demonstrates the practicability of our method.

% \begin{figure}[htbp]
%     \centering
%     \includegraphics[width = 0.29 \textwidth]{fig/k_ablation_vert.pdf}
%     \caption{Target hit ratio on different $k$ (after defense).}
%     \label{fig:ablation}
% \end{figure}

\end{document}